# Investigation of the Zipf-plot of the extinct Meroitic language

*Reginald Smith, Bouchet-Franklin Institute (Decatur, GA)*[1]

**Abstract:** The ancient and extinct language Meroitic is investigated using Zipf's Law. In particular, since Meroitic is still undeciphered, the Zipf law analysis allows us to assess the quality of current texts and possible avenues for future investigation using statistical techniques.

## 1. Introduction

Zipf's law is one of the oldest and most fundamental numerical measures of text structure (Zipf, 1948). There have been many studies of Zipf's Law affirming its universal importance and prevalence among human languages. Rank-frequency distributions of word forms in human languages in most cases abide by the right-truncated Zipf distribution (Zipf 1948; McCowan, et. al. 1999, 2005), however, possessing such a plot is a necessary, but not sufficient, measure of whether the repertoire can represent a real human language. Zipf's work was later expanded and given a firm mathematical footing by connecting it to the burgeoning area of information theory. Both Mandelbrot (1953) and Naranan & Balasubrahmanyan (1992a, b) derived results similar to Zipf's Law by using an entropy based fitness with Mandelbrot using the letter as the primary symbol and Naranan & Balasubrahmanyan using the word as the primary symbol. Structured discourse is not the only possible origin of power law behavior though. Miller (1957) & Li (1992) among others show that a text of random words and spaces can also demonstrate power law behavior. This so-called "monkey language", named for the amusing concept of monkeys randomly typing on typewriters, can produce a Zipfian spectrum with respect to the probability vs. rank of word occurrence. However, Ferrer i Cancho & Solé (2002) have shown that "monkey language" does not present a completely realistic reproduction of human language statistics. In "monkey language", the inverse Zipf plot where the probability of a word having a frequency of appearance in the text P(*f*) is plotted vs. frequency shows non-power law behavior in contrast to the power law behavior of human languages. Thus it seems "monkey language" does not render Zipf's Law a trivial aspect of linguistic statistics. Therefore Zipf's law is still an important and perhaps a fundamental aspect of human languages whose significance is still heavily debated and researched.
    The Zipf plots for many modern languages have been well studied. In this paper, the Zipf plot of the word occurrence frequencies for a corpus of Meroitic texts is investigated in the same light. Since Meroitic has still been undeciphered for almost 100 years, this author believes new techniques borrowed from quantitative linguistics may be useful in illuminating the structure and meaning of the language. As a litmus test for further study, a basic adherence to Zipf's Law can

---

[1] Address correspondence to: Reginald Smith     E-mail: rsmith@sloan.mit.edu

confirm the texts currently in possession have lexical statistics in a manner consistent with other languages and thus may be amenable to other types of mathematical analysis.

## 2. A Short History of Meroitic (Török 1997; Lobban 2004)

Meroitic was the written language of the ancient civilization of Kush, located for centuries in what is now the Northern Sudan. The word 'Meroitic' derives from the name of the city Meroë, which was located on the East bank of the Nile south of where the Atbara River flows off to the east. It is the second oldest written language in Africa after Egyptian hieroglyphs. It is a phonetic language with both a hieroglyph form using some adopted Egyptian hieroglyphs and a cursive form similar to Egyptian Demotic writing. The language had one innovation uncommon in ancient written languages such as Egyptian hieroglyphics or Greek in that there was a word separator, similar in function to spaces in modern scripts, that looks similar to a colon (see Table 1). Meroitic was employed starting the 2nd century BC and was continuously used until the fall of Meroë in the mid 4th century AD.

Table 1
Meroitic Cursive and Hieroglyphic words and their transliterations

[Taken from the latest font set for Meroitic Hieroglyphic and Cursive characters developed by the Meroitic scholars Claude Carrier, Claude Rilly, Aminata Sackho-Autissier, and Olivier Cabon. Web Address: http://www.egypt.edu/etaussi/informatique/meroitique/meroitique01.htm]

| | | | | | | | |
|---|---|---|---|---|---|---|---|
| ᛋ2 | a | ろ | k | /ʔ | q | ♂ | w |
| ⌒ | b | ぅ | l | ω | r | /// | y |
| 2 | d | ɜ | m | 3 | s | : | word separator |
| ʕ | e | ß | n | ⌒// | se | | |
| ⊂ | ẖ | ⋏ | ne | ・ʆ | t | | |
| З | ḫ | / | o | /ʓ̄ | te | | |
| ⅎ | i | ξ | p | ⌒ | to | | |
| 𓀔 | a | 𓅯 | k | △ | q | 𓆈 | w |
| 𓃀 | b | 𓃠 | l | ▫◻ | r | 𓏥 | y |
| 𓂀 | d | 𓃒 | m | 𓅐 | s | ⁞ | word separator |
| 𓃭 | e | 𓈖 | n | ♯ | se | | |
| ◯ | ẖ | ♯♯ | ne | 𓄿 | t | | |
| ᴗ | ḫ | 𓄿 | o | 𓎛 | te | | |
| 𓀀 | i | ⊞ | p | 𓂝 | to | | |

The script was rediscovered in the 19th and 20th centuries as Western archaeologists began investigating the ancient ruins in the Sudan. The first substantial progress in deciphering Meroitic came around 1909 when British archaeologist Francis Llewellyn Griffith was able to use a bark stand which had the names of Meroitic rulers in Meroitic and Egyptian hieroglyphs. The Meroitic hieroglyphs were then corresponded to the Meroitic cursive script and it was then possible to transliterate Meroitic (see Table 1). Some vocabulary was later deciphered by scholars including loan words from Egyptian, gods, names, honorifics, and common items (see Table 2). However, the language remains largely undeciphered. The greatest hope for decipherment, a Rosetta stone type of tablet containing writing in Meroitic and a known language such as Egyptian, Greek, Latin, or Axumite, has yet to be found. Further confounding research is the confusion regarding which language family Meroitic belongs to. Cognate analysis has proceeded extremely slowly since it is disputed to which language family Meroitic properly belongs. Recent work by (Rilly, 2004) has suggested that Meroitic belongs to the North Eastern Sudanic family, however, full decipherment has still proceeded very slowly.

Table 2
Top 20 ranked words and rank-frequency count distributions for REM 1003

| Regular Word | Count | Possible Meaning | Bound Morpheme Removed Word | Count | Possible Meaning |
|---|---|---|---|---|---|
| seb | 10 | ? | li | 25 | particle |
| qoleb | 8 | these?? | seb | 10 | ? |
| qor | 7 | king | qoleb | 8 | these?? |
| tkk | 7 | ? | lw | 8 | same as -lowi (is the, it is)?? |
| amnp | 5 | Amun of Napata | qor | 7 | king |
| abrsel | 4 | Men | tkk | 7 | ? |
| kdisel | 4 | Women | qes | 6 | Kush |
| abr | 3 | Man | amnp | 5 | Amun of Napata |
| adgite | 3 | ? | abrsel | 4 | Men |
| arseli | 3 | ? | kdisel | 4 | Women |
| grpgel | 3 | to command/commander? | lo | 4 | particle('is a/the') |
| grpglke | 3 | to command/commander? | te | 4 | locative particle |
| kdi | 3 | women | abr | 3 | Man |
| mno | 3 | Amun | adgite | 3 | ? |
| ns | 3 | ? | arse | 3 | ? |
| qes | 3 | Kush | grpgel | 3 | to command/commander? |
| qesli | 3 | Kushite? (adj/noun)? | grpglke | 3 | to command/commander? |
| qesto | 3 | Kushite? (adj/noun)? | kdi | 3 | Woman |
| wwikewi | 3 |  | mno | 3 | Amun of Napata |
| 100 (number) | 2 |  | ns | 3 |  |

In light of the slow progress by traditional linguistics methods to translate the language, this author has begun to investigate various methods such as natural language processing and types of statistical analysis to try to gleam more information about the language, its structure, and meaning. This is not an entirely new approach since Meroitic was one of the earliest ancient languages to be investigated using computers (Leclant 1978; Heyler 1970, 1974).

## 3. The Zipf Plot of Meroitic

### 3.1 Mathematical Techniques

The author analyzed the texts by using a computer program to fit them to the Zipf version of the zeta function where the frequency of a word given its rank is given by the following equation:

(1)    $$f_z = \frac{C}{z^\alpha}, \quad z = 1, 2, 3 \ldots n$$

Where $f_z$ is the probability of the word of rank $z$ for ranks 1 to $n$ (right-truncated), $C$ is the normalizing constant, and $\alpha$ is the scaling factor of the power law behavior; $f_z$ is considered zero for all $z$ greater than $n$.

The author analyzed 25 of the longest and most complete texts. These typically have at least 20 different word tokens with a highest frequency count of at least 2. These texts were from the Répertoire d'épigraphie méroïtique (REM), the largest collection of writings in the Meroitic script. Except for a few large stelae, such as those by the Meroitic king Taneyidamani (REM 1044), almost all Meroitic scripts available are funerary texts which follow a similar and well-recognized formulaic pattern for the obituaries recorded within them. The Altmann-Fitter statistical software was used to calculate the parameters and the Zipf distribution and fit it to the data.

Some spellings were standardized since many words were obviously spelled in different ways since there was no standard spelling in Ancient Kush. The author was judicious, however, and did not make specious replacements of similar looking words but words that were used in identical contexts and almost identical spellings. Words which were partially illegible were not corrected and counted as a single instance of an "illegible" word.

In the end, two Zipf distributions were created. The first, is a standard Zipf distribution of the frequency and rank of individual words in the corpus. Individual words were distinguished by being separated by the separator character (which can be found in Table 1). The second Zipf distribution took into account the presence of many conspicuous bound morphemes in the Meroitic language. Many Meroitic verbs, as well, as some nouns have suffixes which contain grammatical meaning. For example, it is known that the suffix *telowi* or *teli* is appended to the name of a place, such as a city, to indicate that the subject of the sentence was affiliated with this place. There is also an extremely common suffix *lowi* or *li* that is appended to nouns that may denote the noun as an indirect object in the sentence. Their definitions are still tenuous, however, these bound morphemes are very common and were separated into independent words for the second Zipf plot. The six bound morphemes separated out were "qo", "lo", "li", "te", "lebkwi", "mhe". They were separated in the manner:

*qo* → *separated out to "qo"*       *lw* → *separated out to "lw"*       *atomhe* → *ato and mhe*
*lo* → *separated out to "lo"*       *telowi* → *te and lo and wi*         *atmhe* → *at and mhe*
*li* → *separated out to "li"*       *teli* → *te and li*                  *qowi* → *qo and wi*
*lowi* → *lo and wi*                 *lebkwi* → *lebk and wi*

### 3.2. Results

The numerical results for each of the individual texts, as well as the entire corpus, is presented here in Table 3. $N$ is the total number of words and $V$ is the number of distinct word types. BM indicates a bound morpheme separated out text.

Table 3
Rank-frequency distribution results for texts

|  |  |  | Zeta Distribution (Zipf) | | | | | Right-Truncated Zeta Distribution | | | | |
|---|---|---|---|---|---|---|---|---|---|---|---|---|
| Text | N | V | a | C | $\chi^2$ | $P(\chi^2)$ | DF | a | C | $X^2$ | $P(\chi^2)$ | DF |
| REM 0088 | 37 | 33 | 0.78 | 0.79 | 29.26 | 0.01 | 13 | 0.32 | 0.03 | 1.01 | 1.0 | 21 |
| REM 0088-BM | 53 | 37 | 0.92 | 0.51 | 27.25 | 0.07 | 18 | 0.64 | 0.09 | 4.70 | 1.0 | 24 |

| REM 0129 | 82 | 71 | 0.61 | 0.62 | 50.98 | 0.04 | 35 | 0.36 | 0.06 | 4.79 | 1.0 | 49 |
|---|---|---|---|---|---|---|---|---|---|---|---|---|
| REM 0129-BM | 118 | 74 | 0.86 | 0.41 | 48.13 | 0.24 | 42 | 0.72 | 0.13 | 15.67 | 1.0 | 51 |
| REM 0217 | 40 | 37 | 0.88 | 0.72 | 28.86 | 0.01 | 14 | 0.32 | 0.06 | 2.44 | 1.0 | 23 |
| REM 0217-BM | 70 | 42 | 0.99 | 0.45 | 31.55 | 0.09 | 22 | 0.81 | 0.16 | 10.96 | 1.0 | 27 |
| REM 0221 | 32 | 25 | 0.91 | 0.52 | 16.78 | 0.08 | 10 | 0.63 | 0.13 | 4.07 | 1.0 | 14 |
| REM 0221-BM | 48 | 29 | 1.09 | 0.48 | 23.23 | 0.06 | 14 | 0.76 | 0.14 | 6.57 | 1.0 | 19 |
| REM 0223 | 23 | 21 | 0.09 | 3.29 | 75.83 | 0 | 2 | 0.29 | 0.02 | 0.56 | 1.0 | 13 |
| REM 0223-BM | 38 | 26 | 0.96 | 0.54 | 20.68 | 0.06 | 12 | 0.66 | 0.05 | 1.74 | 1.0 | 16 |
| REM 0229 | 30 | 29 | 1.08 | 0.91 | 27.21 | 0 | 10 | 0.40 | 0.08 | 2.50 | 1.0 | 16 |
| REM 0229-BM | 49 | 33 | 0.87 | 0.51 | 24.80 | 0.07 | 16 | 0.70 | 0.09 | 4.21 | 1.0 | 21 |
| REM 0237 | 29 | 26 | 0.10 | 2.87 | 83.23 | 0 | 4 | 0.36 | 0.06 | 1.68 | 1.0 | 16 |
| REM 0237-BM | 55 | 32 | 1.05 | 0.33 | 18.10 | 0.32 | 16 | 0.76 | 0.05 | 2.51 | 1.0 | 21 |
| REM 0247 | 45 | 38 | 0.82 | 0.65 | 29.11 | 0.02 | 16 | 0.43 | 0.07 | 3.37 | 1.0 | 25 |
| REM 0247-BM | 59 | 39 | 0.98 | 0.41 | 24.48 | 0.18 | 19 | 0.73 | 0.11 | 6.34 | 1.0 | 25 |
| REM 0259 | 28 | 27 | 0.12 | 2.33 | 65.35 | 0 | 4 | 0.2 | 0.02 | 0.59 | 1.0 | 17 |
| REM 0259-BM | 42 | 31 | 0.86 | 0.51 | 21.44 | 0.09 | 14 | 0.57 | 0.06 | 2.59 | 1.0 | 20 |
| REM 0264 | 33 | 31 | 0.15 | 2.16 | 71.15 | 0 | 6 | 0.23 | 0.02 | 0.73 | 1.0 | 20 |
| REM 0264-BM | 50 | 33 | 0.86 | 0.5 | 24.94 | 0.07 | 16 | 0.62 | 0.04 | 1.98 | 1.0 | 22 |
| REM 0278 | 30 | 30 | 0.21 | 1.82 | 54.60 | 0 | 6 | 0 | 0 | 0 | 0 | N/A |
| REM 0278-BM | 50 | 35 | 0.08 | 4.97 | 248.6 | 0 | 10 | 0.7 | 0.16 | 7.88 | 1.0 | 22 |
| REM 0289 | 48 | 39 | 1.00 | 0.59 | 28.08 | 0.03 | 16 | 0.48 | 0.06 | 2.91 | 1.0 | 25 |
| REM 0289-BM | 65 | 43 | 0.87 | 0.47 | 30.66 | 0.10 | 22 | 0.65 | 0.07 | 4.53 | 1.0 | 29 |
| REM 0297 | 24 | 23 | 0.92 | 0.72 | 17.40 | 0.03 | 8 | N/A | N/A | N/A | 0 | N/A |
| REM 0297-BM | 43 | 27 | 0.94 | 0.45 | 19.25 | 0.12 | 13 | 0.74 | 0.06 | 2.43 | 1.0 | 17 |
| REM 0324 | 27 | 26 | 0.87 | 0.68 | 18.28 | 0.03 | 9 | 0.21 | 0.02 | 0.58 | 1.0 | 16 |
| REM 0324-BM | 45 | 31 | 0.91 | 0.48 | 21.70 | 0.12 | 15 | 0.66 | 0.05 | 2.42 | 1.0 | 20 |
| REM 0386 | 27 | 25 | 0.65 | 0.97 | 26.27 | 0 | 8 | 0.31 | 0.04 | 1.05 | 1.0 | 15 |
| REM 0386-BM | 43 | 28 | 1.08 | 0.46 | 19.64 | 0.10 | 13 | 0.72 | 0.09 | 4.04 | 1.0 | 18 |
| REM 0387 | 31 | 28 | 0.14 | 2.45 | 75.85 | 0 | 5 | 0.27 | 0.02 | 0.76 | 1.0 | 18 |
| REM 0387-BM | 48 | 31 | 1.00 | 0.44 | 21.18 | 0.13 | 15 | 0.66 | 0.05 | 1.59 | 1.0 | 21 |
| REM 0521 | 54 | 50 | 0.14 | 1.51 | 81.57 | 0 | 15 | 0.22 | 0.02 | 1.19 | 1.0 | 35 |
| REM 0521-BM | 80 | 54 | 0.94 | 0.43 | 34.60 | 0.15 | 27 | 0.69 | 0.11 | 9.04 | 1.0 | 36 |
| REM 1003 | 327 | 230 | 0.54 | 0.23 | 74.41 | 1.0 | 155 | 0.46 | 0.03 | 9.11 | 1.0 | 181 |
| REM 1003-BM | 368 | 226 | 0.65 | 0.20 | 72.58 | 1.0 | 157 | 0.59 | 0.04 | 13.52 | 1.0 | 175 |
| REM 1020 | 36 | 31 | 0.99 | 0.66 | 23.65 | 0.02 | 12 | 0.37 | 0.05 | 1.79 | 1.0 | 20 |
| REM 1020-BM | 57 | 36 | 0.99 | 0.41 | 23.34 | 0.18 | 18 | 0.72 | 0.10 | 5.67 | 1.0 | 24 |
| REM 1044 (A-D) | 435 | 362 | 0.45 | 0.21 | 93.35 | 1.0 | 235 | 0.40 | 0.06 | 26.23 | 1.0 | 261 |
| REM 1044(A-D)-BM | 460 | 365 | 0.50 | 0.22 | 99.28 | 1.0 | 241 | 0.46 | 0.08 | 34.74 | 1.0 | 263 |
| REM 1057 | 41 | 38 | 0.97 | 0.99 | 40.41 | 0 | 15 | 0.80 | 0.45 | 18.61 | 0.48 | 19 |
| REM 1057-BM | 61 | 41 | 0.92 | 0.49 | 29.98 | 0.09 | 21 | 0.69 | 0.11 | 6.76 | 1.0 | 27 |
| REM 1064 | 55 | 48 | 0.69 | 0.7 | 38.37 | 0.01 | 21 | 0.28 | 0.02 | 1.36 | 1.0 | 34 |
| REM 1064-BM | 83 | 50 | 0.87 | 0.39 | 32.26 | 0.26 | 28 | 0.7 | 0.05 | 4.33 | 1.0 | 36 |

Table 4 shows the rank frequency distribution for two of the texts used, REM 1003 (the Amanirenas/Akinidad stela) and REM 1020, this information is provided to allow readers to confirm the results with their own analysis. Interestingly, REM 1003 had less word types with the bound morphemes removed since many word types in the regular version appear both alone or with one or more types of suffixes, so separated out you are left with less word types.

Table 4
Rank-Frequency Counts: REM 1020 and REM 1003

**REM 1020**

| **Normal** | | | **Bound Morpheme Removed** | |
|---|---|---|---|---|
| **Rank** | **Frequency** | | **Rank** | **Frequency** |
| 1 | 4 | | 1 | 9 |
| 2 | 2 | | 2 | 8 |
| 3 | 2 | | 3 | 4 |
| 4-31 | 1 | | 4 | 2 |
| | | | 5 | 2 |
| | | | 6 | 2 |
| | | | 7-36 | 1 |

**REM 1003**

| **Normal** | | | **Bound Morpheme Removed** | |
|---|---|---|---|---|
| **Rank** | **Frequency** | | **Rank** | **Frequency** |
| 1 | 10 | | 1 | 25 |
| 2 | 8 | | 2 | 10 |
| 3 | 7 | | 3 | 8 |
| 4 | 7 | | 4 | 8 |
| 5 | 5 | | 5 | 7 |
| 6 | 4 | | 6 | 7 |
| 7 | 4 | | 7 | 6 |
| 8-19 | 3 | | 8 | 5 |
| 20-54 | 2 | | 9 | 4 |
| 55-230 | 1 | | 10 | 4 |
| | | | 11 | 4 |
| | | | 12 | 4 |
| | | | 13 | 4 |
| | | | 14-24 | 3 |
| | | | 25-61 | 2 |
| | | | 62-226 | 1 |

Below are the plots of the data listed. Figure 1 and Figure 2 show the data plots of one of the larges texts, REM 1003 in both normal and bound morpheme removed versions.

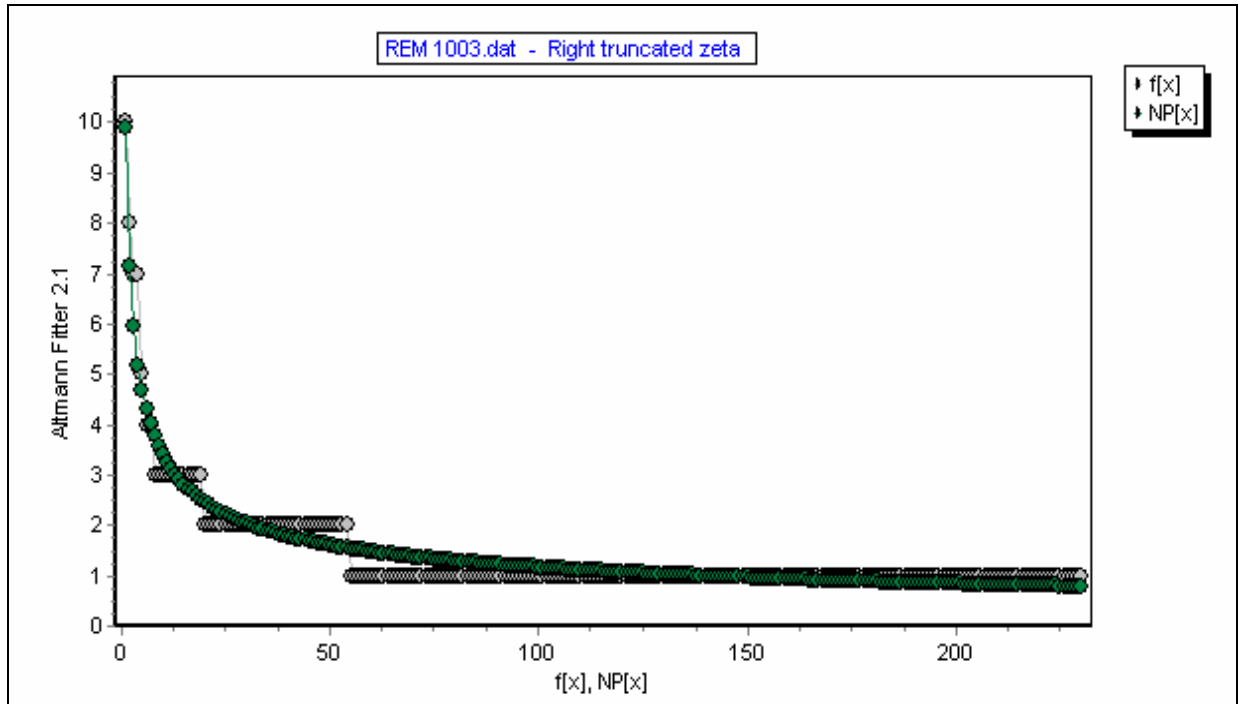

Figure 1. REM 1003 Original Text

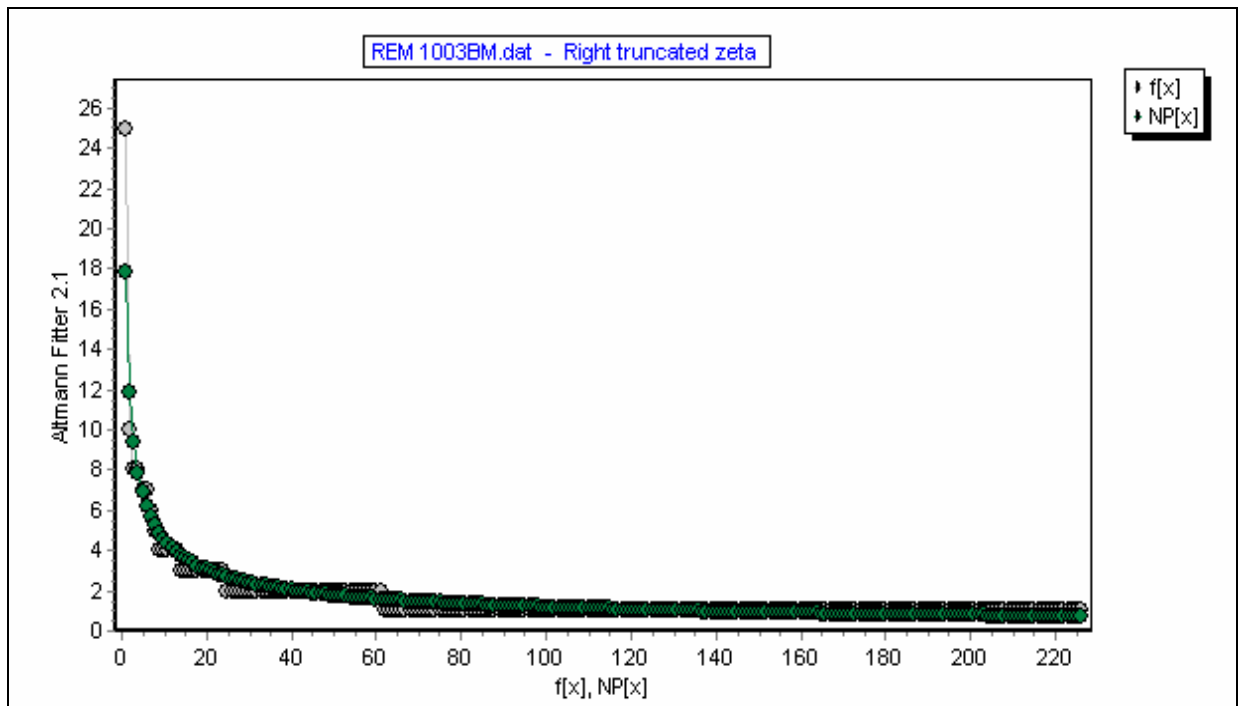

Figure 2. REM 1003 Bound Morpheme Separated

Most texts were funeral texts with a very redundant and standardized structure. The words, *wosi* (Isis)*, soreyi* (Osiris)*, terike* (begotten of)*,* and *tedge* (born of) are present in almost all of the funeral texts and this fact is accounted for in their high ranks. Outside of the funeral texts and their repetitious obituaries, only several long stelae remain which seem to be documentaries of the role of the ruler it is dedicated to (ie. Akinidad, Haramadoye, Taneyidamani, etc.). Given this intuitive result, perhaps the language is more accurately represented with the bound morphemes separated since this populates the necessary high frequency words lacking in the first analysis.

Overall, the Meroitic language shows that it behaves like all other human languages in following the Zipf law for word frequency. This starting point allows us to more confidently use further statistical techniques to further understand Meroitic.

## 4. Conclusion

The above analysis indicates that Meroitic, despite being undeciphered, statistically behaves like all other human languages with a word frequency distribution exhibiting power law behavior. Though our texts are very short, the original requirement of $a \approx 1$ cannot be maintained. If we use the more realistic version of Zipf´s law, namely the truncation at the right side of the distribution, the value of the parameter $a$ is not characteristic of human language but more probably of the given text or genre.

Modern linguistic theory dictates it is unlikely that one can completely reconstruct a natural language based only on statistical data. However, with words that are already known, perhaps one can use context and other methods to infer currently hidden meanings in the Meroitic texts. This paper serves as a foundation for such work by demonstrating that Meroitic behaves statistically like other languages in adhering to Zipf word frequency distributions. Possible future statistical analysis could use some similarity measure to relate unknown words to current known ones and allow us to perhaps infer the meanings of some unknown words by seeing which known words they are similar to. If such an analysis would even be moderately successful, it would open new doors of analysis both in archaeology and mathematical linguistics.

**Acknowledgements**
The author would like to acknowledge both Laurance Doyle and Richard Lobban whose comments were instrumental in finishing this paper.

## References

**Ferrer i Cancho, R., Solé, R.V.** (2002). Zipf's Law and Random Texts. *Advances in Complex Systems 5 (1), 1-6.*
**Heyler, Andre** (1970). Essai de Transcription Analytique des Textes Meroitiques Isoles. *Meroitic Newsletter 5, 4-8.*
**Heyler, Andre (1974).** Meroitic Language and Computers: Problems and Perspectives. In: Abdalla, A.M. (Editor) *Studies in Ancient Languages of the Sudan*: *31-39* Khartoum: Khartoum University Press.
**Leclant, Jean** (1978). The present position in the deciphering of Meroitic script. In: *The peopling of ancient Egypt and the deciphering of Meroitic script : proceedings of the symposium held in Cairo from 28 January to 3 February 1974, 107-119*. Paris: UNESCO.
**Li, Wentian** (1992). Random texts exhibit Zipf's-law-like word frequency distribution. *IEEE Transactions in Information Theory 38 (6), 1842-1845.*
**Lobban, Richard** (2004)**.** *Historical Dictionary of Ancient and Medieval Nubia*. Lanham: Scarecrow.

**Mandelbrot, Benoit** (1953). An informational theory of the statistical structure of language. In: Willis Jackson (ed.), *Communication Theory: 486-502*. London: Butterworths Scientific Publishing.

**McCowan, B., Hanser, S., Doyle L.R.** (1999). Quantitative tools for comparing animal communication systems: information theory applied to bottlenose dolphin whistle repertoires. *Animal Behaviour* 57, 409-419.

**McCowan, B., Doyle, L.R., Jenkins, J.M., Hanser** (2005). The Appropriate Use of Zipf's Law in Animal Communications Studies. *Animal Behaviour* 68, F1-F7.

**Miller, G. A.** (1957). Some effects of intermittent silence. *American Journal of Psychology 70, 311-314*.

**Newman, M.E.J.** (2005). Power laws, Pareto distributions and Zipf's law. *Contemporary Physics* 46, 323-351.

**Rilly, Claude** (2004). The Linguistic Position of Meroitic. *ARKAMANI Sudan Journal of Archaeology and Anthropology*. [http://www.arkamani.org/arkamani-library/meroitic/rilly.htm].

**Török, László** (1997). *The Kingdom of Kush: Handbook of the Napatan-Meroitic Civilization*. London: Brill.

**Zipf, George K.** (1948). *Human behavior and the principle of least effort; an introduction to human ecology*. Cambridge: Addison-Wesley.